\documentclass[letterpaper, 10 pt, conference]{ieeeconf}  

\IEEEoverridecommandlockouts                              

\usepackage{tabularray}
\usepackage{amsmath}
\usepackage{amssymb}
\usepackage{lipsum}
\usepackage{mathtools}
\usepackage{cuted}
\usepackage{algorithm2e}

\usepackage{ragged2e}

\usepackage{tikz}
\usetikzlibrary{shapes,arrows,scopes,positioning,calc,math,patterns,decorations.pathmorphing,decorations.markings,decorations.pathreplacing,spy,angles,quotes}

\usepackage{ctable}
\makeatletter
\def\thickhline{%
  \noalign{\ifnum0=`}\fi\hrule \@height \thickarrayrulewidth \futurelet
   \reserved@a\@xthickhline}
\def\@xthickhline{\ifx\reserved@a\thickhline
               \vskip\doublerulesep
               \vskip-\thickarrayrulewidth
             \fi
      \ifnum0=`{\fi}}

\usepackage{array}

\pretocmd\@bibitem{\color{black}\csname keycolor#1\endcsname}{}{\fail}
\newcommand\citecolor[1]{\@namedef{keycolor#1}{\color{red}}}
\makeatother

\newlength{\thickarrayrulewidth}
\usepackage{multirow}
\usepackage{mcode}

\usepackage[inline]{asymptote}

\usepackage{siunitx}
\usepackage{nicefrac}

\makeatletter

\usepackage{xcolor}
\usepackage[normalem]{ulem}

\usepackage{soul}
\usepackage{lscape}
\usepackage[utf8]{inputenc}
\usepackage{float}
\usepackage{rotating}
\usepackage{pifont}
\newcommand{\cmark}{\ding{51}}%
\newcommand{\xmark}{\ding{55}}%
\usepackage{ragged2e}
\usepackage{hyperref}

\usepackage{makecell}
\usepackage{booktabs}
\usepackage{colortbl}

\begin{document}
\title{\LARGE \bf
A Novel Deep Neural Network for Trajectory Prediction in Automated Vehicles Using Velocity Vector Field 
}

\author{MReza Alipour Sormoli, Amir Samadi, Sajjad Mozaffari, Konstantinos Koufos,\\ Mehrdad Dianati, and Roger Woodman%
\thanks{This research is sponsored by the Centre for Doctoral Training to Advance the Deployment of Future Mobility Technologies (CDT FMT) at the University of Warwick. All authors are with Warwick Manufacturing Group (WMG), The University of Warwick, Coventry CV4 7AL, U.K.
         \{mreza.alipour, amir.samadi, sajjad.mozaffari, konstantinos.koufos, m.dianati, r.woodman\}@warwick.ac.uk}%
}

\maketitle
\thispagestyle{empty}
\pagestyle{empty}
\begin{abstract}
Anticipating the motion of other road users is crucial for automated driving systems (ADS), as it enables safe and informed downstream decision-making and motion planning. Unfortunately, contemporary learning-based approaches for motion prediction exhibit significant performance degradation as the prediction horizon increases or the observation window decreases. This paper proposes a novel technique for trajectory prediction that combines a data-driven learning-based method with a velocity vector field (VVF) generated from a nature-inspired concept, i.e., fluid flow dynamics. In this work, the vector field is incorporated as an additional input to a convolutional-recurrent deep neural network to help predict the most likely future trajectories given a sequence of bird's eye view scene representations. The performance of the proposed model is compared  with state-of-the-art methods on the highD dataset demonstrating that the VVF inclusion improves the prediction accuracy for both short and long-term (5~sec) time horizons. It is also shown that the accuracy remains consistent with decreasing observation windows which alleviates the requirement of a long history of past observations for accurate trajectory prediction.\footnote{Source codes are available at: https://github.com/Amir-Samadi/VVF-TP}.


\end{abstract}

\section{Introduction}
\label{sec: introduction}

Safe and efficient automated driving in dense road traffic environments, where several vehicles dynamically interact with each other, requires predicting their intended manoeuvres and/or future trajectories as a function of time~\cite{lin2016tube}. 
The prediction accuracy becomes essential in this case and directly impacts the downstream motion planning performance of automated driving systems (ADS). On the one hand, a long time prediction horizon (five seconds) allows for a proactive response to the dynamic changes in the driving scene, while, on the other hand, small observation windows (less than one second) are desirable to be able to predict the future states for most of the perceived surrounding vehicles (SVs). To meet both targets, a comprehensive understanding of the spatio-temporal interactions among nearby road users, including semantic environmental information, such as the location of lane markings, road layout, speed limits and nominal velocities is  needed~\cite{liu2019integrated, petrich2013map}. 
Formulating such a complicated trajectory prediction problem for all vehicles in the scene is not viable through human-crafted heuristic algorithms, which explains why data-driven techniques consist the state-of-the-art (SOTA) methods in predicting the motion of other road users for ADS. 
\begin{figure}[t]
\centering
\includegraphics[width=1\linewidth]{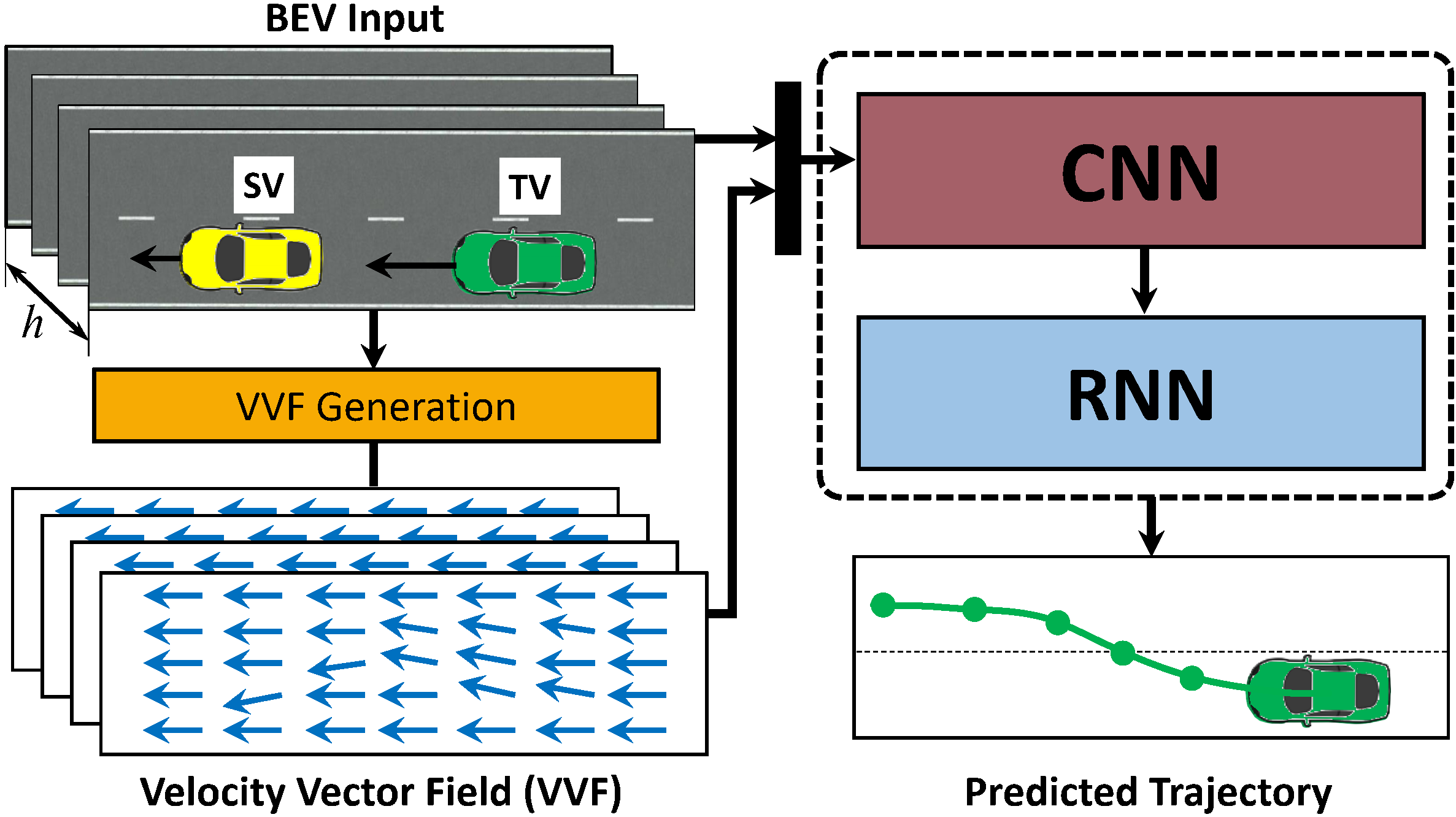}
\caption{General framework of the proposed method operating on BEV maps and incorporating VVF into learning-based methods for trajectory prediction. The trajectory of the TV is predicted while taking into account the interactions among all vehicles including the TV and SVs within an observation window of length $h$.}
\label{fig: framework2}
\centering
\end{figure}

Contemporary learning-based motion prediction methods have widely leveraged convolutional and recurrent techniques to capture the influence of spatial and temporal interactions among road users on their future  trajectories~\cite{deo2018convolutional}. Despite their promising performance in some driving scenarios, these approaches unfortunately face several challenges. Firstly, their prediction accuracy severely degrades as the prediction horizon increases. Secondly, they rely on past state observations within a specific and usually long period of time, which limits their ability to appropriately react to actors that have been only recently detected~\cite{mozaffari2023multimodal}.
\begin{figure*}[t!]
\centering
\vspace{3mm}
\includegraphics[width=0.95\linewidth]{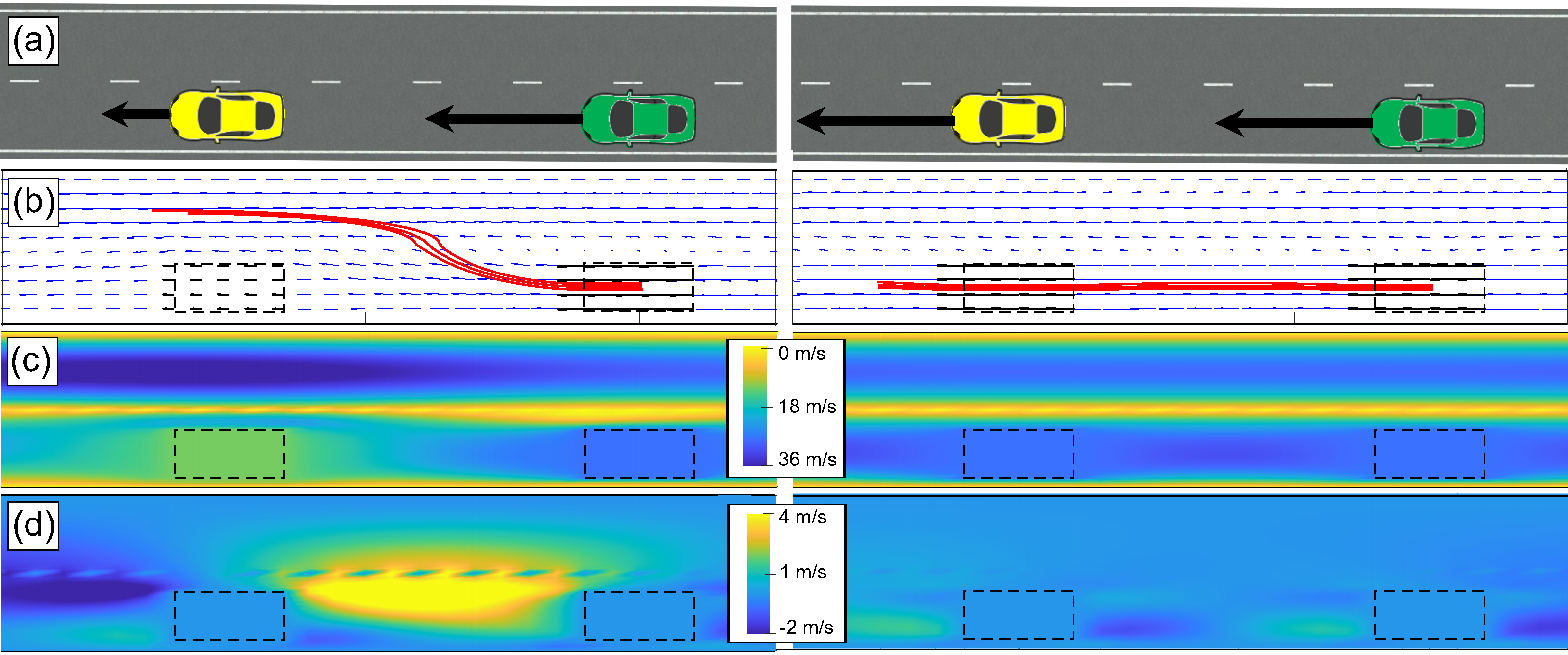}
\caption{An example illustration of a vector field representation for a driving context including a double-lane road and two moving vehicles. In the right and left columns the vehicle in front (yellow) moves at the same speed and at a lower speed than the vehicle behind (green), respectively. (a) Example illustration of the driving scenario. (b) The associated generated vector fields (blue arrows) and three sampled streamlines (an example illustration of predicted trajectories) for the target vehicle (green). (c) and (d) The magnitude of the vector field components in the longitudinal and lateral  directions, respectively.}
\label{fig: VF channels}
\centering
\end{figure*}

In order to address the above shortcomings, this paper proposes a novel hybrid logical-learning method for trajectory prediction, see Fig.~\ref{fig: framework2} for its block diagram representation. Similar to the study in~\cite{li2018humanlike}, a baseline deep neural network (DNN) is designed to predict future trajectories  using a bird's eye view (BEV) map of the driving scene. To better capture the spatio-temporal inter-dependencies among nearby road users, as well as the semantic information of the drivable area, the BEV data is further processed to generate an equivalent velocity vector field (VVF) based on fluid dynamics principles. 
Specifically, each BEV map is associated with a two-dimensional (2d) VVF that provides the most likely speed and orientation of a particle for each pixel of the map.  Therefore, the generated vector field helps distil additional meaningful information from the driving scene and use that to augment the input to the DNN, so that the training quality is enhanced and the prediction accuracy is increased. For instance, Fig.~\ref{fig: VF channels} intuitively illustrates how the VVF information helps distinguish between two different cases in a double-lane driving scenario. The left-hand side scenario includes a low-speed moving vehicle in front of the target vehicle (TV), whereas, both vehicles travel with the same speed in the right-hand side scenario. The streamlines starting from cells occupied by the TV are  derived from the equivalent VVF (Fig.~\ref{fig: VF channels}b), and suggest that the TV is going to carry out a lane change at the left-hand side scenario and lane-keeping at the right-hand side situation. This could not be inferred from traditional BEV inputs in which only the instantaneous locations and velocities of the two vehicles would be available to the DNN. The VVF is essentially a model-based enhancement to the BEV map that allows the DNN to better understand the current and future interactions of vehicles in the scene.  
In summary, the contributions of this paper are:

\begin{itemize}
    \item Introducing a novel approach to represent the spatio-temporal interactions between road users inspired by the fluid dynamics concept. This is achieved by encoding the semantic information of the driving context as a fluid flow simulation and generating a VVF accordingly.
    \item Developing a novel trajectory prediction method that combines a convolutional-recurrent DNN with a continuous VVF. This technique results in a notable improvement in trajectory prediction accuracy ranging from 18\% to 72\% compared to SOTA methods. 
    \item A VVF dataset associated with the publicly available highD dataset, which can be leveraged by the wider research community for designing sophisticated data-driven learning-based methods for trajectory prediction, decision-making, motion planning and control for ADS. 
\end{itemize}

The rest of this paper is organised as follows: After reviewing  related studies in Section~\ref{section: related_works}, the problem, system model assumptions and proposed method are explained in Section~ \ref{section: method}. The performance evaluation of the proposed method is compared with SOTA results in Section~\ref{sec: result}. Finally, the key takeaways of this study and highlights for future work are presented in Section~\ref{sec: conlusion}.

\section{Related work}
\label{section: related_works}
A review of recent learning-based vehicle trajectory prediction methods has been provided in~\cite{mozaffari2020deep}. This section reviews existing studies on vehicle trajectory prediction, focusing on two main aspects: (1) The input representation used in learning-based prediction, and (2) the type of deep learning techniques used for vehicle trajectory prediction.

\subsection{Input Representations for Trajectory Prediction}
The future trajectory of a target vehicle (TV)  depends on several factors such as its current and previous states, the interaction with surrounding vehicles (SVs) and the scene context. Early studies only encoded the TV's motion states into the prediction model's input~\cite{xin2018intention, park2018sequence} yielding accurate short-term predictions, but failing to predict the TV's trajectory for longer time horizons. To overcome this issue, recent  studies also encoded the interaction with SVs as a list of features, such as the relative distance/velocity between the TV and every SV in the scene~\cite{altche2017lstm, deo2018multi, diehl2019graph}. 
It is also possible to augment the feature list with some scene context features such as the existence of adjacent lanes or the lane width~\cite{mozaffari2023multimodal}. As expected, the main drawback of these methods lies in the implementation complexity of the learning model. 
Similarly, end-to-end approaches that automatically extract features from raw-sensor data~\cite{luo2018fast, casas2018intentnet}, also suffer from high computational complexity. Kim 
 {\it{et al.}}~\cite{kim2021lapred} addressed this issue by introducing trajectory-lane features, where the scene context  and vehicle features are jointly learnt per driving lane. Another group of studies utilised a simplified BEV representation of the environment, including the scene context and the dynamic agents within the environment~\cite{deo2018convolutional, itkina2019dynamic, mozaffari2022early}. Deo {\it{et al.}} in~\cite{deo2018convolutional} utilised a social grid map where each occupied cell by a vehicle is filled with the encoded vehicle dynamics using a long short-term memory (LSTM) network. In~\cite{itkina2019dynamic}, a dynamic occupancy grid is generated from ego-centric point cloud data. Mozaffari {\it{et al.}} in~\cite{mozaffari2020deep} utilised a stack of BEV images, each representing the dynamic and static context of the scene at a specific time step. Similarly, this paper adopts a BEV representation for the input data, however, we enrich the BEV images with the generated VVF of the driving environment. While traditional BEV input representations are usually sparse, unless the road traffic density is extremely high, the proposed VVF augments the BEV input with extra information that helps deduce the TV's future speed at future locations. 

\subsection{Deep Learning Techniques for Trajectory Prediction}
Recurrent neural networks, more specifically LSTMs, have been used in several studies for vehicle trajectory prediction~\cite{altche2017lstm, xin2018intention, park2018sequence}. LSTMs take advantage of feedback loops for extracting temporal features from sequential input data. Despite their power in learning the features of sequential data, they lack a mechanism for spatial features extraction, which is also required for understanding the spatial interactions for vehicle trajectory prediction problems. Therefore, several studies utilised deep learning techniques, such as attention mechanisms~\cite{messaoud2020attention}, and convolutional neural networks (CNNs)~\cite{deo2018convolutional, mukherjee2020interacting}, often as an addition to LSTMs, to fill this gap. Messaoud {\it{et al.}} in~\cite{messaoud2020attention} proposed a multi-head attention pooling mechanism to extract spatial information from encoded dynamics of vehicles, while in~\cite{deo2018convolutional}, CNNs are used to serve the same task. Mukherjee et al. in~\cite{mukherjee2020interacting} utilised convLSTM to extract spatio-temporal features from a sequence of occupancy grid maps of the driving environment. Similarly, in this paper, we first use a CNN to extract spatial features from a sequence of VVF-augmented BEV input representation. The results are then processed by an LSTM to extract temporal dependencies. The comparative study shows that the proposed method can outperform existing SOTA approaches.


\section{Proposed Framework}
\label{section: method}



This section describes the VVF-based Trajectory Prediction \textbf{(VVF-TP)} framework. It is first discussed how the BEV input data is processed to  generate the associated VVF and how the latter is fed into the input of the DNN. Afterwards, the DNN pipeline including a joint convolutional-recurrent neural network is presented and the learning procedure is subsequently described.
\begin{figure*}[t]
\centering
\vspace{3mm}
\includegraphics[width=1\linewidth]{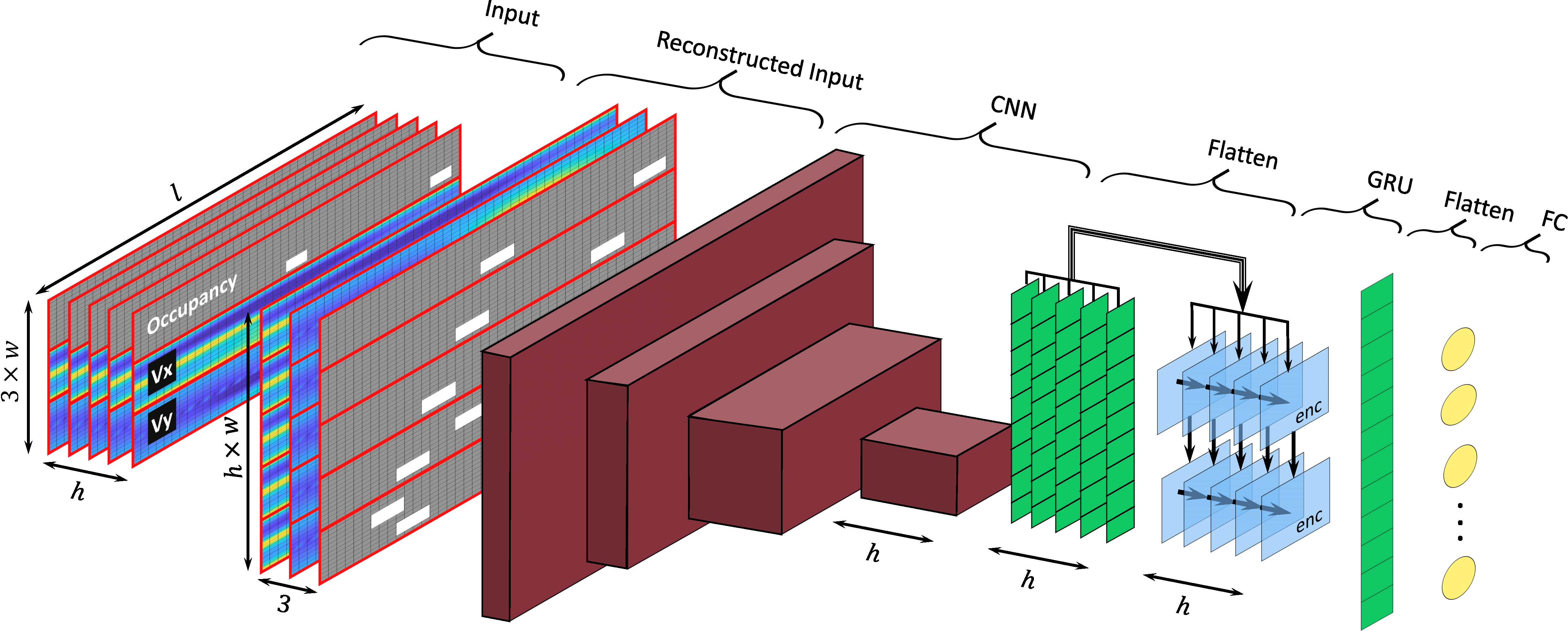}
\caption{Structure of the proposed DNN for trajectory prediction. The input comprises the occupancy grid, the longitudinal, $V_x$, and lateral, $V_y$, components (their magnitude) of the generated VVF over the observation length $h$. The length and width of the BEV images are $l$ and $w$, respectively.}
\label{fig: framework}
\centering
\end{figure*}

\subsection{VVF Generation}
\label{subsection: model}
A boundary-value fluid flow dynamic problem is employed for encoding the semantic information of the driving context as a continuous vector field over a 2d drivable area. 
First, we explain how this problem is derived, and next, how it is solved using a numerical method. 

\subsubsection{Formulating fluid flow problem from driving context}
The boundary-value problem is defined using fluid flow simulation in a structured channel that has the same geometry with the equivalent drivable area. In our case, four types of semantic information of the driving context are interpreted into boundary conditions (BCs) of the fluid flow simulation, namely the moving vehicles' velocities, road boundaries, lane markings, and nominal lane speeds.  To consider the \textbf{\textit{moving vehicles' velocities}}, the fluid particle's velocity at the cells occupied by vehicles is set equal to the current vehicle speeds. Since the \textbf{\textit{road boundary}}, similar to static obstacles, should be avoided, a \textit{no-slip} condition is associated with it, whereas the \textbf{\textit{lane markings}} should be passable to allow for lane-change manoeuvres. Accordingly, the \textit{porous} condition is applied to the cells corresponding to lane markings so that the fluid particles could pass the porous barrier in spite of its resistance. Finally, the input/output velocity at the fluid channel cross sections is set equal to the \textbf{\textit{nominal speed}} (longitudinal) per lane, e.g., 30~m/sec for  highways. After the values of the vector field  at the boundaries are set, the remaining values are determined  by solving the underlying  Partial Differential Equations (PDE) that are known as Navier-Stokes equations~ \cite{glowinski2003numerical}. 

\subsubsection{Solving the fluid flow problem}
While there is no analytical solution for the boundary-value problem, numerical methods, proposed in the computational fluid dynamics literature, can be used to solve for the VVF of the BEV map. Among all available solutions, the lattice Boltzmann method (LBM)~\cite{chen1998lattice} is adopted, as it can handle BCs with sophisticated geometries, and it is also easily parallelizable. Both features make this method suitable for motion prediction applications in ADS, where computational efficiency and digesting complex driving contexts are key requirements. 

In LBM the simulated fluid domain is discretised into uniformly spaced grids on a lattice (grid cells for the 2d domain in our case study). After assigning the BCs to the corresponding grids, the LBM  solves the fluid dynamics differential equation indirectly and calculates the motion vectors via two main steps, i.e., propagation (streaming) and collision (relaxation) of fluid density in the lattice. There are different ways of connecting adjacent nodes in a lattice which are called lattice vectors. For instance, the D2Q9 lattice vector means that the lattice is a 2d grid and each node is connected to nine surrounding nodes. At each lattice update, the microscopic density is propagated along the lattice vector and the node densities are updated through the collision process (see the Appendix for more details). Fig.~\ref{fig: VF channels}b illustrates the VVF calculated for the  different scenarios depicted in Fig.~\ref{fig: VF channels}a. The corresponding longitudinal, Fig.~\ref{fig: VF channels}c, and lateral, Fig.~\ref{fig: VF channels}d, velocities that satisfy the BCs are also depicted. 
\begin{table}
\centering
\caption{Network parameters for different layers, input, and output. The observation length and the predicted trajectory length are $h$ and $p$, respectively.}
\begin{tblr}{
  column{1} = {1cm},
  column{1} = {1.8cm},
  column{3} = {3.5cm},
  row{1-15} = {c,m},
  cells = {c},
  cell{2}{1} = {r=2}{},
  cell{4}{1} = {r=2}{},
  cell{6}{1} = {r=3}{},
  cell{10}{1} = {r=8}{},
  hline{1,9} = {1.3pt},
  hline{2,10,18} = {0.8pt},
  hline{4,6} = {0.3pt},
  hline{3,5,7,8,11-17} = {0.3pt, dashed},
}
\textbf{Layer} & \textbf{Type} & {\textbf{Output Shape}}         \\
Input          & Initial       & $(h,3 \times 32,256) $                              \\
               & Reconstructed & $(3,h \times 32,256) $                              \\
GRU            & Flatten       & $(h,\text{CNN}_{feature \_ size})$                           \\
               & Recurrent     & $(h,64)$                                            \\
FC             & Flatten       & ($h \times 64) $                                    \\
               & F$_x$         & $(p)$                                               \\
               & F$_y$         & $(p)$                                               \\
\textbf{Layer} & \textbf{Type} & {\textbf{Configuration}\\ (Filter, Kernel, Stride)} \\
CNN            & ReLu(Conv2D)  & $(32,3,1)$                                          \\
               & MaxPool2D     & $(-,2,2)$                                           \\
               & ReLu(Conv2D)  & $(32,3,1)$                                          \\
               & ReLu(Conv2D)  & $(32,3,1)$                                          \\
               & MaxPool2D     & $(32,3,1)$                                          \\
               & ReLu(Conv2D)  & $(32,3,1)$                                          \\
               & MaxPool2D     & $(32,3,1)$                                          \\
               & ReLu(Conv2D)  & $(h,3,1) $                                          
\end{tblr}
\label{table: model_arch}
\end{table}
The lattice-based propagation and collision processes make it possible to run the LBM on parallel architectures such as graphics processing units (GPUs) that enable real-time performance. \textit{Sailfish}~\cite{januszewski2014sailfish} is a well-developed open-source toolbox that implements the LBM  with flexibility for defining a problem and incredible computational performance on GPU up to 400 million lattice updates per second (mlups) on GeForce RTX 3080 Ti. Therefore, if it takes 100 lattice updates to calculate the VVF for a lattice with $256\times 64$ dimensions, the computation takes only $4.4$~ms to complete. Note that the nominal transmission frequency of the Collective Perception Messages (CPM) standardised by ETSI is equal to $10$~Hz. It is therefore expected that the EV will receive the perception obtained by roadside infrastructures equipped with sensors and wireless connectivity every $100$~ms.
 
\subsection{BEV occupancy grid and VVF data representation}
\label{subsec: data_rep}
 With the generated VVF at hand, this section explains how the occupancy grid map and the VVF are pre-processed before being fed into the DNN, see the first two steps at the left-hand side of Fig.~\ref{fig: framework}: {\it{Input}} and {\it{Reconstructed Input}}. The occupancy grid consists of pixels (or cells) and represents the BEV image using ternary values for each pixel, specifically, (1) for unoccupied cells, (2) for cells occupied by road participants, and (3) for  cells covered by the TV.  The VVF is represented by  two float-valued images, with sizes equal to that of the occupancy grid map, which contain the calculated longitudinal and lateral velocity for each cell of the map. At the {\it{Input}} step, one may observe that for each element of the observation window $h$, we have vertically stacked the  occupancy grid and the VVF, where $w$ and $l$ denote the number of pixels representing the width and length of the BEV map, respectively. At the {\it{Reconstructed Input}} data preparation step, the input data is  transformed into three spatio-temporal images, where we have concatenated the observation windows of each image. To the best of our knowledge, for the first time in the related literature, we introduce grid time-sequence images in the data preparation step, enabling the following CNN layers to obtain a more comprehensive overview of the entire observation time. 

The size of the BEV occupancy grid is configured as $32 \times 256$ pixels.  The 20-meter lateral coverage has enough capacity for the TV's adjacent participant to be included, and the 200-meter longitudinal coverage assures enough space for comprising a complete lane change or overtaking manoeuvre. The pre-processed input data to the DNN is a three-channel image of size ($h \times 32 \times 256$) representing BEV, longitudinal and lateral VVF.

\subsection{GRU-CNN Model Design}
\label{subsec: DNN}
This section presents the DNN model architecture that is employed to generate trajectories for the TV using the pre-processed historical observation data of BEV images and VVFs. Fig.~\ref{fig: framework} illustrates the model architecture, wherein the initial layers comprise CNN layers succeeded by a Recurrent Neural Network (RNN) encoder. CNNs are widely acknowledged for extracting image features, and enabling the identification of spatial patterns through convolution and pooling operations. Each convolutional layer applies a set of filters to the input image, producing a collection of feature maps that capture distinct aspects of the BEV and VVF input images. Subsequently, the pooling layers downsample the feature maps while preserving important information. These layers are succeeded by Gate Recurrent Unit (GRU) layers, which capture temporal features extracted by the CNN. GRU layers, similar to LSTM networks extensively employed in the literature \cite{deo2018convolutional, messaoud2019non}, possess fewer parameters to train and facilitate learning. The GRU layers consist of gated units that selectively retain or forget information from preceding time steps, enabling the network to capture sequential information. Afterwards, to decode the extracted spatio-temporal features we used fully connected (FC) layers, transferring encoded feature size into the predicted trajectory for the TV. 
The specific architecture and hyper-parameters of the CNN layers utilised in this study are presented in Table~\ref{table: model_arch}.

It is worth noting that in the existing literature, it is common to employ a decoder RNN subsequent to the encoder GRU. However, in our study, we deliberately omitted this layer to compel the CNN to comprehend both spatial and temporal aspects of the provided observation history. This is accomplished by aligning the output channel number of the final CNN layer and the encoding GRU layer with the observation history $h$. Considering the high computational costs associated with RNN layers, our study demonstrates that avoiding an additional decoder GRU layer not only enhances the overall agility of the architecture but also encourages the CNN layers to extract spatio-temporal features.

\subsection{Training Process}
During the training phase, the DNN estimates the future trajectory of the TV and compares it with its ground truth by evaluating the (differentiable) ``Huber Loss" function, $L_\delta$, which applies a $\delta$ threshold to strike a balance between the Mean Squared Error (MSE) and the Mean Absolute Error (MAE). Let us define by $(x_j,y_j), j=1,\ldots p$ the coordinates of the ground truth trajectory on the BEV map, and by $(\hat{x}_j,\hat{y}_j)$ the predicted trajectory on the same coordinate system. Let us also construct the $2p$-dimensional column vectors of stacked coordinates, i.e., $\mathbf{z}$ for the ground truth and $\mathbf{\hat{z}}$ for the predicted trajectory. Then, the Huber loss function can be read as: 
\begin{equation}
\label{eq: loss}
\mathcal L_{\delta} = \begin{cases}
\frac{1}{2} ||\mathbf{\hat{z}} -\mathbf{z} ||_2^2, &  || \mathbf{\hat{z}} -\mathbf{z}||_1 \leq\delta. \\
\delta \left(|| \mathbf{\hat{z}} -\mathbf{z}||_1-\frac{\delta}{2}\right),  & \textnormal{otherwise}.
\end{cases}
\end{equation}

While the quadratic nature of MSE magnifies the values of outliers to avoid them, the MAE weights all errors larger than $\delta$ on the same linear scale. In practice, our focus lies on improving the overall trajectory prediction performance rather than solely mitigating outlier errors. To this end, intuitively, the Huber Loss function applies MSE to small error values, which results in amplifying them, and MAE to large error values, leading the DNN to spend less learning time to avoid outlier errors.

\section{Performance Evaluation}
\label{sec: result}

The performance evaluation of the trajectory prediction method developed in this article (hereafter referred to as the VVF-TP) is carried out for highway driving scenarios in two parts. First, a comparative study to assess the performance against the SOTA  methods, second, an ablation analysis to show how the VVF and observation window length affect the prediction performance, separately. In this section, the selected dataset and the evaluation metrics are explained, before presenting the quantitative/qualitative performance evaluation results. During training the parameter $\delta$ in the Huber Loss function in Eq.~(1) is set at $\delta=1$~m, and the rest of network parameters are given in Table.~\ref{table: model_arch}.
\def\ra{.12}
\def\w{1.2}
\renewcommand{\arraystretch}{1.7}
\begin{table}
\centering
\vspace{3mm}
\caption{Comparing the trajectory prediction error in meters based on Eq.~\eqref{eq: metrics} for various methods. The best and second-best results are in bold and underlined, respectively.}
\begin{tabular}{>{}m{2 cm}c>{\centering}m{\w cm}c>{\centering}m{\w cm}c>{\centering}m{\w cm}c>{\centering}m{\w cm}c>{\centering}m{\w cm}} 
\toprule
\textbf{Methods} & \textbf{1s} & \textbf{2s} & \textbf{3s} & \textbf{4s} & \textbf{5s} \\
\midrule
CS-LSTM~\cite{deo2018convolutional} & \underline{0.19} & 0.57 & 1.16 & 1.96 & 2.96 \\
\hline
STDAN~\cite{chen2022intention} & \underline{0.19} & \underline{0.27} & \underline{0.48} & 0.91 & 1.66 \\
\hline
MMNTP~\cite{mozaffari2023multimodal} & \underline{0.19} & 0.36 & 0.56 & \underline{0.82} &\underline{1.19}\\
\hline
S-LSTM~\cite{coifman2017critical} & 0.22 & 0.62 & 1.27 & 2.15 & 3.41 \\
\hline
NLS-LSTM~\cite{messaoud2019non} & 0.22 & 0.61 & 1.24 & 2.10 & 3.27 \\
\hline
VVF-TP & \textbf{0.12} & \textbf{0.24} & \textbf{0.41} & \textbf{0.66} & \textbf{0.98} \\

\bottomrule
\end{tabular}
\label{table: result}
\end{table}

\def\wb{.54}
\begin{table*}
\centering
\vspace{3mm}
\caption{Comparing the effect of four different input states on the RMSE of the prediction error. Column $x$ corresponds to the RMSE of the longitudinal distance between the two trajectories, Eq.~\eqref{eq: metrics_x}, column $y$ corresponds to the RMSE of the lateral distance and column $R$ to the RMSE of the distance, Eq.~\eqref{eq: metrics}. The best and second-best results are in bold and underlined, respectively.}
\begin{tabular}{>{\centering}m{1 cm}c>{\centering}m{\wb cm}c>{\centering}m{\wb cm}c>{\centering}m{\wb cm}c>{\centering}m{\wb cm}c>{\centering}m{\wb cm}c>{\centering}m{\wb cm}c>{\centering}m{\wb cm}c>{\centering}m{\wb cm}c>{\centering}m{\wb cm}c>{\centering}m{\wb cm}c>{\centering}m{\wb cm}c>{\centering}m{\wb cm}c>{\centering}m{\wb cm}c>{\centering}m{\wb cm}c>{\centering}m{\wb cm}c>{\centering}m{\wb cm}c>{\centering}m{\wb cm}}
\toprule
\multirow{2}{1 cm}{\textbf{State }} & \multirow{2}{\wb cm}{\textbf{VVF}} & \multirow{2}{\wb cm}{\textbf{Long Obs}} &    &    \textbf{1s}      &               &   &    \textbf{2s }     &               &   &   \textbf{3s }   &               &  &     \textbf{4s }    &               &     &     \textbf{5s }    &               \\
\cline{4-18}
                &              &                    & x             & y             & \cellcolor{lightgray} R             & x             & y             & \cellcolor{lightgray}R             & x             & y             & \cellcolor{lightgray} R             & x             & y             &\cellcolor{lightgray} R             & x             & y             &\cellcolor{lightgray} R             \\       
\hline
\textbf{S1}      &      \xmark        &        \xmark            & 1.53          & 0.10          & \cellcolor{lightgray} 1.53          & 2.84          & 0.19          & \cellcolor{lightgray} 2.85          & 4.14          & 0.27          & \cellcolor{lightgray} 4.15          & 5.44          & 0.34          &\cellcolor{lightgray} 5.45          & 6.73          & 0.41          & \cellcolor{lightgray} 6.75          \\
\textbf{S2}      &        \xmark      &         \cmark      & 0.28          & \underline{0.07}          & \cellcolor{lightgray} 0.30          & 0.57          & \underline{0.14}          & \cellcolor{lightgray} 0.60          & 0.93         & \underline{0.20}         & \cellcolor{lightgray}0.95          & 1.36          & \underline{0.27}         &\cellcolor{lightgray} \underline{1.39}          & 1.86          & \underline{0.33}          & \cellcolor{lightgray}1.89          \\
\textbf{S3}      &     \cmark         &           \xmark         & \underline{0.25}          & 0.10          & \cellcolor{lightgray} \underline{0.27}          & \underline{0.48}          & 0.18          & \cellcolor{lightgray} \underline{0.50}          & \underline{0.76}         & 0.25         & \cellcolor{lightgray}\underline{0.80}          & \underline{1.12}          & 0.32         &\cellcolor{lightgray} 1.62          & \underline{1.56}          & 0.38          & \cellcolor{lightgray}\underline{1.60}          \\
\textbf{S4}      &      \cmark      &       \cmark        & \textbf{0.17} & \textbf{0.05} & \cellcolor{lightgray} \textbf{0.12} & \textbf{0.22} & \textbf{0.10} & \cellcolor{lightgray}\textbf{0.24} & \textbf{0.39} & \textbf{0.15} & \cellcolor{lightgray}\textbf{0.41} & \textbf{0.63} & \textbf{0.21} & \cellcolor{lightgray}\textbf{0.66} & \textbf{0.94} & \textbf{0.28} &\cellcolor{lightgray} \textbf{0.98} \\
\bottomrule
\end{tabular}
\label{table: result2}
\end{table*}

\begin{figure*}[t]
\centering
\includegraphics[width=1\linewidth]{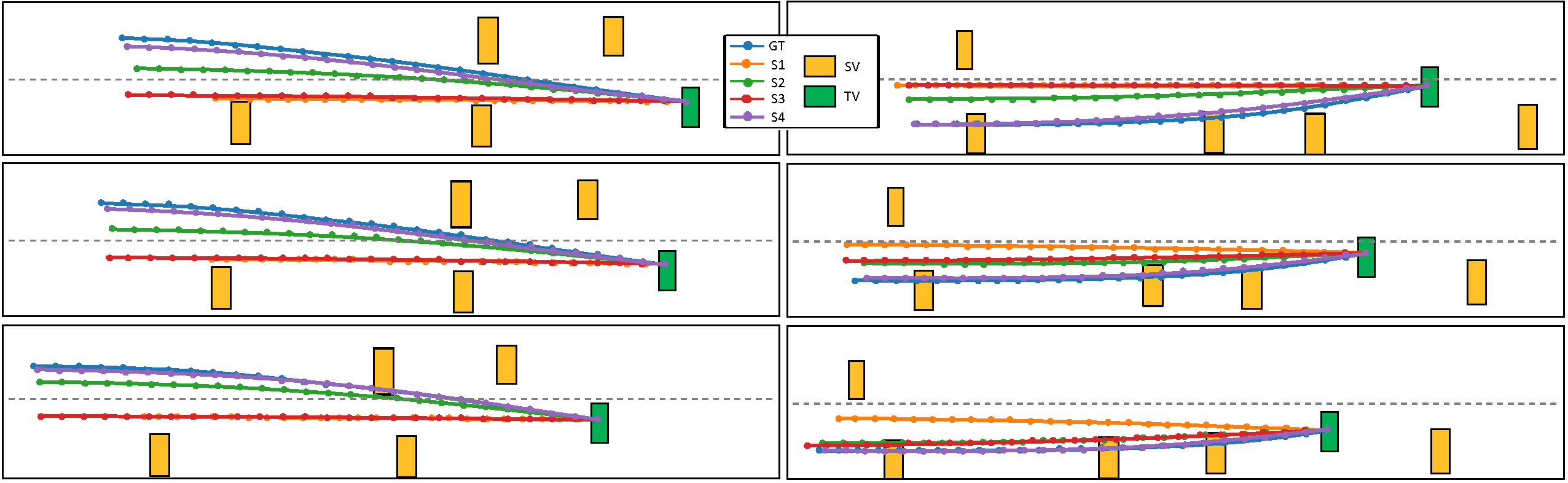}
\caption{Comparing the predicted trajectories using different input states for left and right lane-change (right and left column, respectively) for three consecutive frames (starting from top to bottom) along with the ground truth (GT). The target vehicle (TV) and the surrounding vehicles (SVs) are coloured green and amber respectively.}
\label{fig: qualitative}
\centering
\end{figure*}

\subsection{Dataset}
The highD dataset has been collected in six different locations in Germany and contains 110,000 unique trajectories for various types of vehicles moving on two or three-lane roads~\cite{krajewski2018highd}. It has been widely used for evaluating trajectory prediction performance in highway driving scenarios, as it covers both heavy and light traffic conditions, which cause different driving behaviours throughout the dataset. For a fair comparison of the final results with the SOTA  literature, the highD dataset has been divided into training, testing, and validation subsets with ratios 70-20-10 \%, respectively. The time granularity between consecutive frames is $0.2$~seconds. It is worth mentioning that the associated VVF to this dataset using the methodology described in Section~III.A has been made publicly available at \href{https://github.com/Amir-Samadi/VVF-TP}{https://github.com/Amir-Samadi/VVF-TP}. 

\subsection{Evaluation Metrics}
Similar to several recent studies, the root mean square error (RMSE) of the distance between the predicted trajectory and its ground truth has been used in this paper to measure the prediction accuracy (the lower the RMSE the better) of the VVF-TP method. We report the RMSE separately for the longitudinal ($x$) and the lateral ($y$) directions, in addition to the  RMSE of the distance separation between the two trajectories that are usually reported in the literature. This allows us to obtain a more in-depth understanding of how lateral/longitudinal accuracy contributes to the overall prediction error. The RMSE of the distance between the predicted trajectory and its ground truth can be written as 
\begin{equation}
    \label{eq: metrics}
    \begin{array}{l}
{\rm{RMSE}} = \sqrt {\frac{1}{p}\sum\limits_{j = 1}^p \left[\left(\hat{x}_j - x_j \right)^2 + \left( \hat{y}_j - y_j \right)^2 \right]},
\end{array}
\end{equation}
while the RMSE, e.g., at the longitudinal direction, is calculated as 
\begin{equation}
    \label{eq: metrics_x}
    \begin{array}{l}
{\rm{RMSE}} = \sqrt {\frac{1}{p}\sum\limits_{j = 1}^p {{{\left( {\hat{x}_j - {x_{j}}} \right)}^2}}},
\end{array}
\end{equation}
and a similar expression can be written for the lateral error. 

\subsection{Comparative Results}
There are several SOTA methods for trajectory prediction in which the evaluation metrics are reported based on the highD dataset. These methods can be divided into: (i) Single-modal prediction models such as the CS-LSTM~\cite{deo2018convolutional}, S-LSTM~\cite{coifman2017critical}, and  NLS-LSTM~\cite{messaoud2019non} that are developed based on social pooling and LSTM, and (ii) multi-modal approaches like the STDAN~\cite{chen2022intention} and MMNTP~\cite{mozaffari2023multimodal} that consider more than one possible manoeuvres. The performance evaluation using the RMSE calculated in Eq.~\eqref{eq: metrics} is presented in Table~\ref{table: result}  for five timesteps in the future, starting from one to five seconds with step equal to a second. The VVF-TP outperforms all other methods for all timesteps. The RMSE is reduced (on average) by 16~\%  in all five timesteps with respect to the second-best result that belongs to the recently published works in~\cite{mozaffari2023multimodal, chen2022intention}. 

\subsection{Ablation Analysis}
In this section, the impact of  augmenting the input data to the DNN using the generated VVF and the effect of the size of the  observation window (history of past observations for the target vehicle) on the quality of trajectory prediction are investigated in detail. Bearing that in mind, the prediction accuracy using the RMSE calculated in Eq.~\eqref{eq: metrics} and Eq.~\eqref{eq: metrics_x} is evaluated for four different input states $S1$ to $S4$. In $S1$, the observation window degenerates to the  current perception frame, i.e., $h=1$ and the VVF is omitted from the input. The input state $S2$ is the same as $S1$ except that an observation window of $h=10$~frames (or two seconds) is available for the target vehicle. In $S3$, the VVF is considered with $h=1$ frame and $S4$ takes into account the VVF and $h=10$ frames. 

The quantitative and qualitative results for the four states described in the previous paragraph are given in Table~\ref{table: result2} and Fig.~\ref{fig: qualitative}, respectively. According to Table~\ref{table: result2}, the second-best performance in the longitudinal direction belongs to $S3$, whereas $S2$ performs better in the lateral direction. The overall prediction performance based on the RMSE in Eq.~\eqref{eq: metrics} is better in $S3$ than $S2$ (in four timesteps out of five) indicating that the inclusion of the VVF in the input data is more beneficial than a long history of past observations. Next, according to Fig.~\ref{fig: qualitative}, $S4$ predicts much earlier the lane change trajectory than the other input configurations suggesting that the combination of the long observation window with the inclusion of the VVF lead to the best performance. Moreover, similar to Table~\ref{table: result2}, Fig.~\ref{fig: qualitative} illustrates that the $S3$ incurs larger/smaller prediction errors in the lateral/longitudinal direction than $S2$. Finally, adding a VVF when there is only a single observation available can significantly improve the  longitudinal prediction accuracy (compare 
$S3$ with $S1$ in Fig.~\ref{fig: qualitative}-left).



\section{Conclusion}
\label{sec: conlusion}

The results obtained in this study suggest that a nature-inspired phenomenon such as fluid flow motion has the potential to be used for modelling the imminent interactions among road participants, and add meaningful information in the form of a velocity vector field (VVF) to the conventional bird's eye view representation of the driving scene. Accordingly, a learning-based trajectory prediction approach adapted to digest the new information could achieve higher performance than the state-of-art methods in short/long-term prediction horizons. Also, the model's performance remains consistent when the number of available past states of surrounding vehicles decreases, which would prove useful in driving environments with occlusions.

The proposed prediction model in this study was designed and tested for highway driving scenarios, however, the same logic could apply to more generic conditions such as urban environments. Adapting the proposed model to operate in complex road segments such as urban intersections and roundabouts is an important area of future work. Moreover, regarding practical concerns, the VVF generation process should operate in real-time to become a part of decision-making, motion planning and control of  an autonomous vehicle. These limitations should be addressed in a future study to prove the applicability of the proposed approach in real-life operations. 


\appendix
\label{app: lbm}

\begin{figure}[t!]
\centering
\vspace{3mm}
\includegraphics[width=0.95\linewidth]{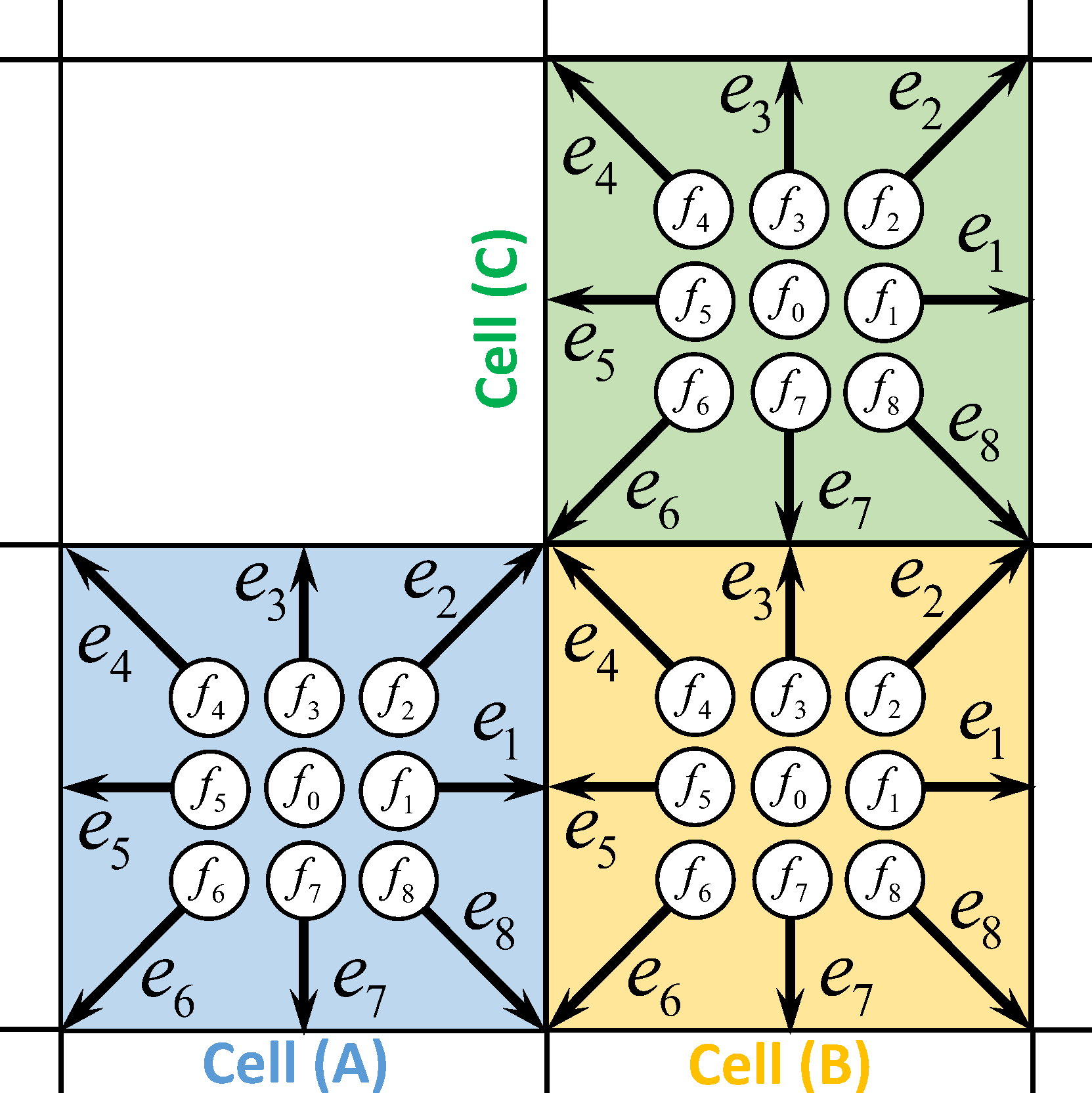}
\caption{Adjacent lattice cells interaction in the overall D2Q9 grid.}
\label{fig: lbm}
\centering
\end{figure}

To generate the 2d velocity vector field using the Lattice Boltzmann Method (LBM) with (D2Q9), each cell in the lattice (the BEV map) interacts with eight surrounding cells plus itself (nine directions) via 2d lattice vectors of unit magnitude,  ${\vec e_i}$, see Fig.~\ref{fig: lbm}. Firstly, the velocity vectors at the boundary cells are set; Recall from Section~III.A the four boundary conditions used in the fluid flow simulation. Then, the velocity vector, $\vec u$, for each of the remaining cells is initialized, and based on that a density  $f_i, i=0,\ldots 8$ is assigned to each direction $e_i$. The values of $(f_i, \vec u)$ for each cell are iteratively updated until there is no significant change in the 
 average velocity ($<~0.01$~m/s) for all cells and the boundary conditions including speed limits are also satisfied. Each iteration consists of the following two steps:

 \textbf{Streaming step:} Adjacent cells exchange the densities between opposite vectors. For instance, in Fig.~\ref{fig: lbm} cell A interacts with cells B and C by exchanging density via $\left\{ {f_5^A \leftarrow f_5^B,f_1^A \to f_1^B} \right\}$ and $\left\{ {f_6^A \leftarrow f_6^C,f_2^A \to f_2^C} \right\}$ pairs, respectively.
 
\textbf{Collision step:} This is executed in each cell separately. First, the equilibrium density is calculated using nine densities and their contributing weights ($\omega_i$). Subsequently, the nine densities and the cell's velocity vector are updated using Eq.~\ref{eq: collision} below (order matters). 
\begin{equation}
\label{eq: collision}
{\begin{array}{*{20}{l}}
{f_i^{eq} = {\omega _i}\rho \left[ {1 + 3\left( {{{\vec e}_i} \cdot \vec u} \right) - \frac{3}{2}\left( {\vec u \cdot \vec u} \right) + \frac{9}{2}{{\left( {{{\vec e}_i} \cdot \vec u} \right)}^2}} \right]},\\
{\;\;{f_i} = {f_i} + {{\left( {f_i^{eq} - {f_i}} \right)} \mathord{\left/
 {\vphantom {{\left( {f_i^{eq} - {f_i}} \right)} \tau }} \right.
 \kern-\nulldelimiterspace} \tau }},\\
{\;\;\,\,\rho  = \sum\limits_{i = 0}^8 {{f_i}} },\\
{\;\;\,\,\vec u = \sum\limits_{i = 0}^8 {{f_i}{{\vec e}_i}} }.
\end{array}}
\end{equation}

According to~\cite{januszewski2014sailfish}, the updating weights for stationary ($\vec e_0$), diagonal ($\vec e_i, i\in \{2,4,6,8\}$), and orthogonal ($\vec e_i, i\in \{1,3,5,7\}$) directions are $\frac{4}{{9}}$, $\frac{1}{{36}}$ and $\frac{1}{{9}}$, respectively. Finally, in fluid flow simulations, $\tau$ is the update rate obtained from the kinematic viscosity property of the fluid which has been set equal to $0.003$ in this study. It should be noted that at each iteration update, the velocity vectors corresponding to the boundary conditions are set equal to the boundary values. 

\bibliographystyle{myIEEEtran}
\bibliography{IEEEabrv,bibliography}
\end{document}